\documentclass[letterpaper]{article} % DO NOT CHANGE THIS
\usepackage{aaai24}  % DO NOT CHANGE THIS
\usepackage{times}  % DO NOT CHANGE THIS
\usepackage{helvet}  % DO NOT CHANGE THIS
\usepackage{courier}  % DO NOT CHANGE THIS
\usepackage[hyphens]{url}  % DO NOT CHANGE THIS
\usepackage{graphicx} % DO NOT CHANGE THIS
\usepackage{natbib}  % DO NOT CHANGE THIS AND DO NOT ADD ANY OPTIONS TO IT
\usepackage{caption} % DO NOT CHANGE THIS AND DO NOT ADD ANY OPTIONS TO IT
\usepackage{algorithm}
\usepackage{algorithmic}

\usepackage{newfloat}
\usepackage{listings}
\DeclareCaptionStyle{ruled}{labelfont=normalfont,labelsep=colon,strut=off} % DO NOT CHANGE THIS
\floatstyle{ruled}
\newfloat{listing}{tb}{lst}{}
\floatname{listing}{Listing}
\usepackage{multirow}
\usepackage{listings}
\usepackage{array}
\usepackage{amsmath}
\usepackage{amssymb}
\usepackage{xcolor}
\usepackage{xspace}
\newcolumntype{P}[1]{>{\centering\arraybackslash}p{#1}}

\newcommand{\methodNameLong}{BridgeQA}

\usepackage{tabularx}
\newcolumntype{Y}{>{\centering\arraybackslash}X}
\title{Bridging the Gap between 2D and 3D Visual Question Answering: A Fusion Approach for 3D VQA}

\author {
    Wentao Mo\textsuperscript{\rm 1},
    Yang Liu\textsuperscript{\rm 1,2}\thanks{Corresponding author}
}
\affiliations {
    \textsuperscript{\rm 1}Wangxuan Institute of Computer Technology, Peking University\\
    \textsuperscript{\rm 2}National Key Laboratory of General Artificial Intelligence, Peking University \\
    mowt@pku.edu.cn, yangliu@pku.edu.cn
}

\usepackage{bibentry}
\begin{document}

\maketitle

\begin{abstract}
In 3D Visual Question Answering (3D VQA), the scarcity of fully annotated data and limited visual content diversity hampers the generalization to novel scenes and 3D concepts (e.g., only around 800 scenes are utilized in ScanQA and SQA dataset).
Current approaches resort supplement 3D reasoning with 2D information. However, these methods face challenges: either they use top-down 2D views that introduce overly complex and sometimes question-irrelevant visual clues, or they rely on globally aggregated scene/image-level representations from 2D VLMs, losing the fine-grained vision-language correlations.
To overcome these limitations, our approach utilizes question-conditional 2D view selection procedure, pinpointing semantically relevant 2D inputs for crucial visual clues. We then integrate this 2D knowledge into the 3D-VQA system via a two-branch Transformer structure. This structure, featuring a Twin-Transformer design, compactly combines 2D and 3D modalities and captures fine-grained correlations between modalities, allowing them mutually augmenting each other. 
Integrating proposed mechanisms above, we present \methodNameLong{}, that
offers a fresh perspective on multi-modal transformer-based architectures for 3D-VQA. Experiments validate that \methodNameLong{} achieves state-of-the-art on 3D-VQA datasets and significantly outperforms existing solutions.
Code is available at \url{https://github.com/matthewdm0816/BridgeQA}.
\end{abstract}

\section{Introduction}
\begin{figure}[!ht]
    \centering
    \includegraphics[width=\columnwidth]{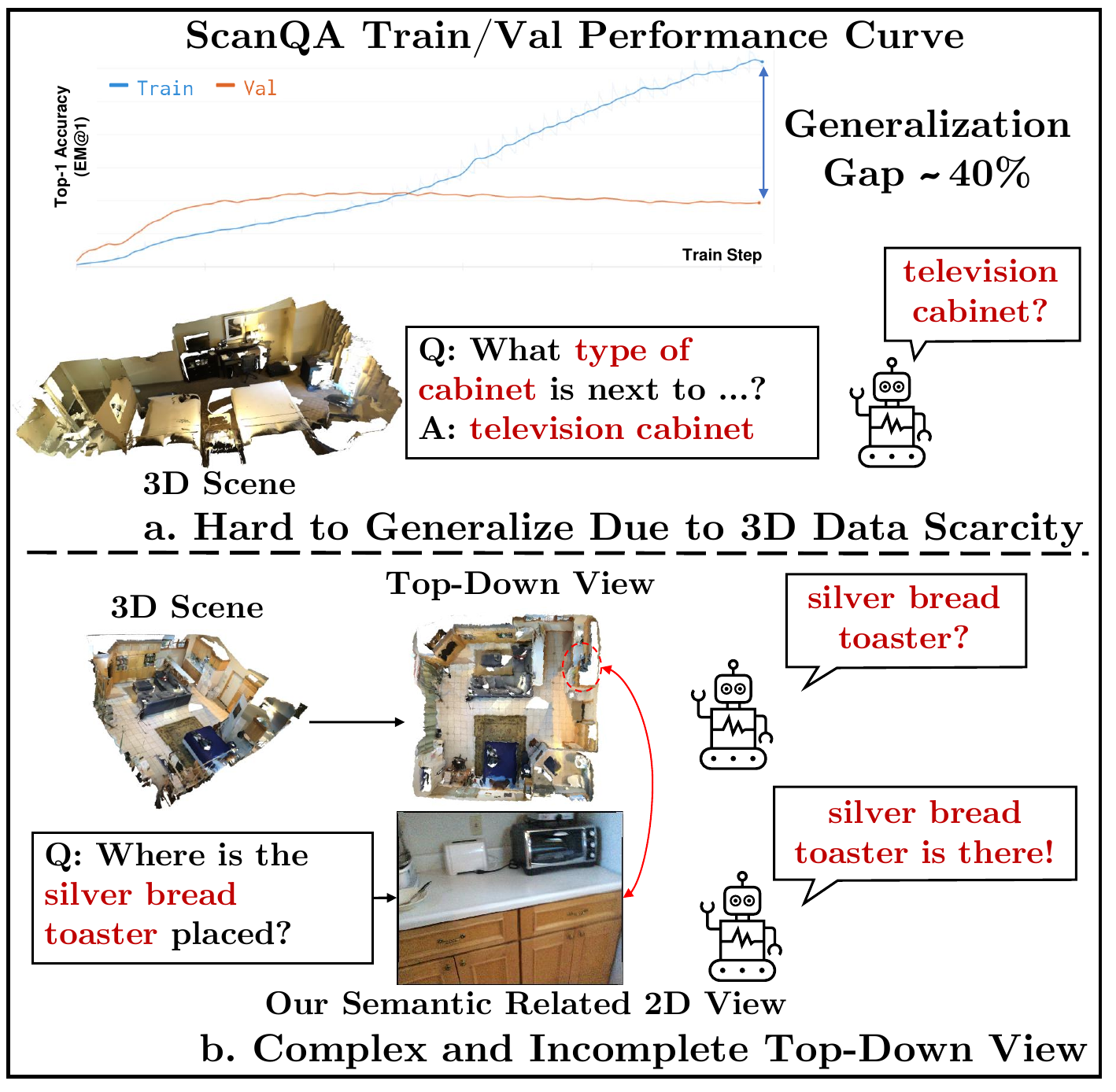}
    \caption{Caveats of current 3D-VQA methods.
    (a). Current 3D-VQA methods exhibits large generalization gap due to 3D data scarcity. 
    And there are questions with visual concepts that never appeared (e.g., ``television cabinet") in either question, answer or object type annotation during training, which are hard for 3D-VQA models to generalize to.
    (b). Current 3D-VQA methods that explictly incorporate 2D VLMs use top-down views of 3D scenes, which might be too complex with many irrelative visual clues and might be incomplete on relative visual clues for some questions. In \methodNameLong{}, we use question-related views instead, to capture visual context potentially relative to the question.
    }
    \label{teaser}
\end{figure}

The pursuit of better vision-language (VL) understanding in AI has been relentless, with extensive research aimed at enabling models to see, read, and reason in response to multi-modal inputs. This is particularly evident in the field of visual question answering (VQA), where models are trained to interpret and respond to visual contents and questions \cite{li2020oscar,tan2019lxmert,yu2019deep,anderson2018bottom}. However, the majority of VQA researches have focused on 2D visual inputs such as images or videos, leaving a gap in the understanding of real-world 3D scenes.

Recognizing the above limitation on understanding 3D inputs, 3D visual question answering (3D-VQA) has emerged as a novel and essential type of 3D scene understanding task. Recent proposals in ScanQA \cite{azuma2022scanqa} and SQA \cite{ma2022sqa3d} have gained increasing research interest, requiring the VQA model to comprehend the 3D scene rather than planar pixels. Most 3D-VQA methods rely on a combination of a 3D object detector and a Transformer-based vision-language fusion backbone to predict an answer.

Despite these advancements in incorporating 3D modality in VQA, the current performance of 3D-VQA tasks remains constrained, particularly in generalizing to novel 3D visual concepts. As illustrated in Figure \ref{teaser}(a), a large generalization gap between training and inference exists for current 3D-VQA method (namely ScanQA \cite{azuma2022scanqa}). Some visual concepts never appeared in question, answer or object type annotations in training split, making it hard for 3D-VQA method to generalize.
This limitation is largely attributed to the scarcity of visual content in existing datasets.  In popular 3D-VQA datasets like ScanQA and SQA, only about 800 indoor scenes are used to formulate questions. However, such challenge could potentially be addressed by incorporating 2D vision-language models familiar with diverse visual concepts.

To leverage both 2D and 3D visual clues, some previous 3D-VQA works \cite{parelli2023clip,delitzas2023multi} have employed the CLIP \cite{radford2021learning} model to enhance 3D scene representation through scene-level feature alignment loss using the 2D image of a top-down view. However, two problems arise in these approaches: 1) the top-down view's complexity and incompleteness
hinders the normal 2D vision-language model's understanding of the 2D view, limiting the infusion of 2D knowledge into the model. As illustrated in Figure \ref{teaser}(b), there can be too much visual context in the top-down image irrelative with the question, while too little information about the asked object occuluded under top-down view. 2) by using image/scene-level rather than region/object-level representations, this global-level modality alignment approach ignores fine-grained 2D-3D correspondence. And by only utilizing the computed feature and ignoring fine-grained 2D VL relation captured within the 2D vision-language model (VLM), they inefficiently used the pretrained 2D VLM.

In response to these challenges, we propose \methodNameLong{}, a novel 3D-VQA approach that incorporates 2D vision-language models into the 3D-VQA framework, aiming to bridge the gap between 2D and 3D VQA. Our model comprises two parts:
1) we propose a \textbf{quesiton-conditional 2D view selection} mechanism, to select scene views semantically related to the question for the input of 2D branch to efficiently infuse 2D information. In this way, the most related 2D visual clues could be captured (as illustrated in Figure \ref{teaser}b) , rather than too complex/incomplete top-down views. 
2) we propose a \textbf{two-branch Transformer-based model} architecture that accepts both 2D and 3D input, implementing a simple yet efficient 2D-3D fusion strategy that compactly incorporate 2D and 3D representations. This two-branch architecture retains as much knowledge from 2D vision-language pretraining as possible. To further align the 2D and 3D modalities, we introduce \textbf{Twin-Transformer}, a method to mix 2D and 3D visual tokens (such as 2D image patches or 3D objects) for two Transformers by cross-attention. This connection allows for implicit alignment with fine-grained correspondence between the 2D and 3D modalities, and makes 2D/3D visual tokens a mutual augmentation for each other.  
We evaluate our proposed method on two popular 3D-VQA datasets, namely ScanQA \cite{azuma2022scanqa} and SQA \cite{ma2022sqa3d}, and significantly surpassed previous second-best methods by 4.3\% and 7.8\%  on two test splits of ScanQA, and 4.4\% on SQA, achieving state-of-the-art performance on both datasets.

Concretely, we make the following contributions: 
1) We present a novel 2D-3D VQA approach that uniquely bridges the gap between 2D and 3D VQA models, addressing the challenges of data scarcity and limited visual content diversity.    
2) We innovate a question-conditional 2D view selection method that identifies semantically relevant 2D input, providing a more nuanced understanding of the visual content.
3) We design a two-branch Transformer model that efficiently fuses 2D and 3D input, employing Twin-Transformer, a simple yet efficient way to mix 2D and 3D tokens while preserving pretrained 2D knowledge, which enhances the robustness and generalizability of 3D VQA systems. 
4) We evaluated our method on two popular 3D-VQA datasets, ScanQA and SQA and significantly surpassed previous state-of-the-art methods, with a comprehensive ablation study confirming our design choices.

\section{Related Work}
\label{sec:rw}

\textbf{Visual Question Answering (VQA):}
Visual Question Answering has emerged as a prominent research area in recent years. Classical approaches often encode images and questions using pre-trained networks, subsequently fusing them to predict an answer. Attention-based and pretraining-based methods have been particularly successful in this domain, achieving significant success in 2D-VQA benchmarks \cite{zhang2021vinvl, yu2019deep, li2020oscar, wang2021simvlm}, even surpassing human performance in some benchmarks \cite{wang2021simvlm}. However, these methods primarily focus on 2D images or videos, lacking the ability to understand 3D modalities. Simply adopting 2D-VQA structures for 3D-VQA is suboptimal as revealed in \cite{azuma2022scanqa, ma2022sqa3d}, which may due to the data scarcity in 3D scenes. In our method, we propose \methodNameLong{} to bridge the gap between 2D and 3D VQA by introducing a two-branch Transformer architecture that accepts both modalities. 

\noindent \textbf{3D VQA:}
While there has been initial researches on 3D-VQA datasets and methods \cite{azuma2022scanqa, ma2022sqa3d, parelli2023clip, delitzas2023multi, zhao2022towards}, the exploration of efficient understanding of 3D scenes and answering visual questions remains limited. Existing 3D-VQA models often resemble their 2D counterparts in structure, comprising a 3D vision encoder instead of 2D encoder with a 3D object detector, a question text encoder, and an attention-based vision-language fusion encoder for cross-modal feature extraction. This approach, however, neglects valuable 2D information within the scenes. 
Some researches \cite{parelli2023clip, delitzas2023multi} have attempted to incorporate 2D information in 3D-VQA by utilizing image-level 2D features from top-down views of 3D scenes. However, this combination is often coarse, and the top-down perspective lacks many essential details. To address above challenges, our method introduces a two-branch Transformer architecture, \methodNameLong{}, that captures both 2D and 3D visual contexts by incorporating a 2D branch into the 3D VQA model. For the 2D input of the VQA model, our method strategically selects question-relative 2D views automatically rather than top-down views that are too complex including many irrelavant objects, enabling the capture of a more comprehensive range of relevant 2D visual concepts.

\noindent \textbf{Pretraining for VQA:} 
Pretraining vision-language models (VLMs) for VQA has been proven to be highly effective, with external knowledge from large-scale paired image-text enhancing the finetuned model's ability substantially \cite{li2020oscar, tan2019lxmert, chen2020uniter}. This approach has been instrumental in understanding diverse 2D visual concepts and enhancing 2D VQA performance. However, the limited availability of 3D scene-text pair data makes large-scale pretraining for 3D VQA very difficult. 
Recent 3D
pretraining methods either propose augmented scene-text datasets \cite{zhu20233dvista}, or propose complicated pretraining tasks \cite{Jin_2023_CVPR} \cite{zhang2023visionlanguage}, and these methods are still confined due to the limited 3D scene diversity (e.g., less than 1200 scenes for all these methods, compared to more than 2M image-text pairs in CC3M \cite{sharma2018conceptual} used for 2D vision-language pretraining). Moreover, without pretraining, simple encoding of 3D scenes struggles to generalize to novel and rare visual concepts due to the scarcity of 3D scenes (e.g., only about 800 scenes in both ScanQA and SQA, compared to more than 200K images in 2D VQA dataset VQA 2.0 \cite{gokhale2020vqa}). In our method, we propose to leverage 2D vision-language pretrained model that could understand diverse visual concepts within the 3D scenes, exploting the 2D pretraining for 3D VQA, to reduce the need for 3D vision-language data.

\section{Method}
\label{sec:method}

\begin{figure*}[h!]
    \centering
    \includegraphics[width=0.9\textwidth]{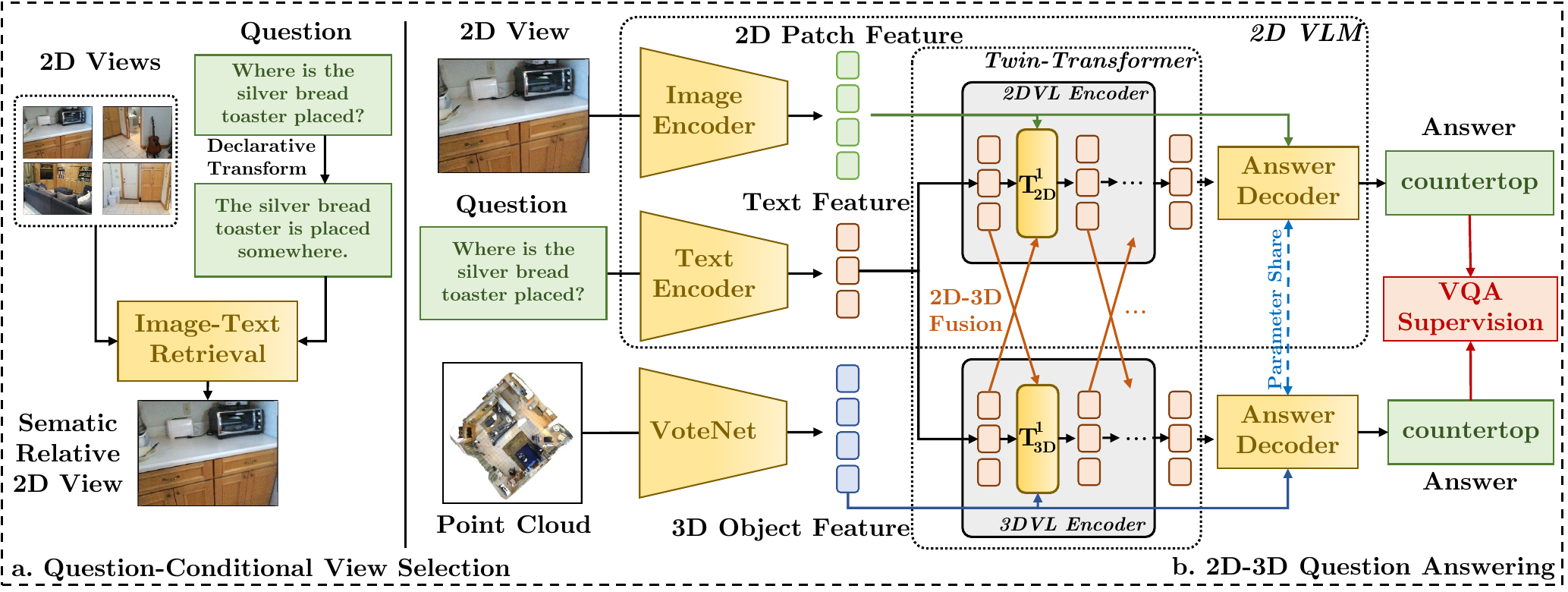}
    \caption{Overview of \methodNameLong{}. 
    (a): In question-conditional view selection, we identify semantic-related 2D views by retrieving images that align with the question's declaration form. This method captures relevant visual cues to enhance the 2D-3D question answering model.
    (b): Our 2D-3D VQA framework utilizes a Twin-Transformer structure, comprising two branches: a 2D vision-language model (VLM) and a 3D branch of similar structure. We apply a lightweight 2D-3D fusion operation. This integration infuses 2D visual context into 3D VQA without modifying the underlying 2D VLM architecture, preserving the pre-trained 2D VL knowledge while allowing for compact fusion of intermediate representations.
    }
    \label{fig:method}
\end{figure*}

% \subsection{Problem Formulation} 
\subsection{Method Overview}  
The 3D VQA task necessitates answering visual questions about 3D scenes, utilizing datasets comprising 3D scene, question, and answer triplets $\mathcal D=\{(\mathcal S_i, q_i, a_i)\}, i=1...N$ 
with $N$ samples. Our 3D-VQA model is required to predict the answer $a$ based on the 3D scene $\mathcal S$ and the question $q$, in an open-ended generative manner.

Our method, \methodNameLong{}, consists of two main components: \textbf{question-conditional 2D view selection} and the \textbf{2D-3D VQA model}, as shown in Figure \ref{fig:method}. For a data sample $(\mathcal S, q, a)$ of 3D scene, question and answer triplet, we first select a \textbf{relevant 2D view} $\mathcal I$ matching the question semantics with an image retrieval model. We could potentially capture more relative visual concepts from selected views than top-down scene view or random views, which allows fine-grained interaction between view and question later.
Then, the novel \textbf{2D-3D VQA model} seamlessly integrates 2D images and 3D scenes through a two-branch transformer initialized with a pretrained 2D VLM. Our \textbf{Twin-Transformer} architecture further ensures that the intermediate features from both modalities are compactly fused, employing a cross-attention mechanism for implicit alignment. With the two-branch structure and proposed Twin-Transformer architecture, our method aligns 2D and 3D visual contexts without compromising the pretrained 2D VL knowledge. Finally, the pretrained 2D VLM's language decoder generates free-form answers, enhancing the model's ability to respond with long text if required, thereby offering a robust and nuanced solution to 3D VQA.

\subsection{Question-Conditional 2D View Selection}

In the realm of 3D Visual Question Answering (3D-VQA), previous methods \cite{delitzas2023multi, parelli2023clip} have attempted to utilize 2D input and often relied on top-down images. While this approach may seem intuitive, it presents challenges as top-down images are not commonly used in 2D vision models, which often lack the necessary context to answer the question or includes too much irrelative details. 
In our method, we address this challenge by selecting 2D views that are most semantically related to the questions. This approach aims to effectively incorporate needed visual concepts for the question, while filtering off irrelatvie visual clues, to improve the model's ability to interpret and respond to 3D scenes. 
This method involves using BLIP's image-text retrieval model to select the most suitable 2D views from video frames, guided by either questions or question-based declarative texts. The question-based declarations are generated merely from question using GPT-4 \cite{openai2023gpt4}. 
During training and inference, both declarations or questions can be used to select 2D views. This flexibility doesn't significantly affect the results, as our empirical studies have shown in Table \ref{tab:main}. More results are shown in ablation (Table \ref{tab:ablation-selection}).

\subsection{2D-3D VQA Model}

The \methodNameLong{} model for VQA consists of two parts, as depicted in Figure \ref{fig:method}(b). The two-branch 2D and 3D VL encoding backbone encodes 2D images and 3D scenes into patch-wise and object-wise visual tokens, respectively, and fuses them with the question text to acquire 2D-language and 3D-language features. The two branches adopt structure and initial parameters from 2D VLM, to exploit 2D VL knowledge from 2D pretraining. 
The 3D branch is initialized from 2D VLM, as it might reduce domain shift between 2D and 3D features between two branches, bringing advantage in 2D-3D feature alignment.
The question-answering decoder leverages the auto-regressive capability of the pretrained language model to generate free-form answers using the fused VL features.

\noindent \textbf{Two-branch 2D and 3D Vision-Language Encoding:}
With semantically related 2D views, we propose a 2D-3D VQA model using a two-branch transformer to accept both 2D images and 3D scenes and compactly utilize the 2D VLM.
For each data sample $(\mathcal S, \mathcal I, q, a)$, consisting of the 3D scene, related 2D image, question, and answer, we first extract 3D object-level features $h_{3D}$ using a 3D object detector, as conventionally adopted in previous 3D-VQA methods. Simultaneously, we extract 2D patch-level features $h_{2D}$ using the visual encoder in a 2D VLM
\begin{equation}
    \begin{aligned}
        h_{2D} &= \operatorname{Transformer}_{VLM,I}(\mathcal I) \\
        h_{3D} &= \operatorname{OD}_{VoteNet}(\mathcal S) \\
    \end{aligned}
\end{equation}
where $\operatorname{Transformer}_{VLM,I}$ is the VLM's image encoder, extracting patch-wise feature tokens $h_{2D}$, and $\operatorname{OD}_{VoteNet}$ is the VoteNet \cite{qi2019deep} 3D object detector backbone, extracting object-wise feature tokens $h_{3D}$. The text feature is extracted by the VLM's text encoder to encode the text feature $h_{q}$.
\begin{equation}
    \begin{aligned}
        h_{q} &= \operatorname{Transformer}_{VLM,q}(q) \\
    \end{aligned}
\end{equation}
where $\operatorname{Transformer}_{VLM,q}$ is the VLM's text encoder.

Previous methods align image-level 2D features with scene-level representations, thereby dismissing fine-grained relations between image regions and scene objects. For example, globally pooled scene-level representation from all objects may lose detailed semantics describing specific object attributes (e.g., red apple), but focusing on more high-level semantics (e.g., a living room). These methods also disposed the pretrained 2D VLM structure and weights, thereby losing valuable 2D VL relations learned from 2D pretraining.
In contrast, our approach maximizes the utilization of 2D pretraining to recognize diverse visual concepts. To fully exploit the capabilities of the 2D pretrained VLM, we opt to directly utilize the aligned VL fusion encoder and language decoder from the 2D VLM, rather than relying on pooled scene or image-level representations. This approach allows us to fuse the 2D feature tokens $h_{2D}$ and 3D object tokens $h_{3D}$ with the text feature $h_{q}$ through a dedicated 2D/3D VL transformer
\begin{equation}
    \begin{aligned}
        h'_{2D} &= \operatorname{Transformer}_{VLM, 2D}(h_{q}, h_{2D}) \\
        h'_{3D} &= \operatorname{Transformer}_{VLM, 3D}(h_{q}, h_{3D}) 
    \end{aligned}
    \label{eqn:2d}
\end{equation}
% \mwt{put back 2D 3D from X}
where $\operatorname{Transformer}_{VLM, 2D}, \operatorname{Transformer}_{VLM, 3D}$ are the 2D/3D VL fusion encoders of VLM, fusing the features into joint VL feature $h'_{2D}$ or $h'_{3D}$. The 3D branch shares the same structure with the 2D branch and is initialized with the other branch's parameters.

\noindent \textbf{Twin-Transformer:}
While a simple two-branch transformer structure can capture both 2D and 3D visual inputs, it may allow only limited interaction between the 2D and 3D visual contexts. However, arbitrarily adding extra modules into the 2D VLM based encoder might harm the 2D VL knowledge from pretraining. 
Thus, a lightweight but compact modality fusion mechanism between 2D and 3D branch might be crucial to the successful knowledge transfer and intergration from 2D to 3D.
To address these challenges, we propose a Twin-Transformer structure that that mix 2D/3D intermediate VL tokens with 3D/2D 
visual tokens used in cross-attention in each branch's Transformer backbone, allowing compact 2D-3D interaction without altering the basic model architecture. 
Moreover, in this way, each visual modality's branch could perceive the complementary visual clues from the other modality.

Specifically, after each layer of the two transformers, the cross-attention input on one branch is augmented with intermediate representations from the other branch, thereby fostering a more nuanced integration of 2D and 3D visual information without altering the basic model structure. Formally, we represent the 2D or 3D backbone $\operatorname{Transformer}_{VLM,X}$ with $L$ layers of transformer blocks $(T^1_{X},...,T^L_{X})$ for modality $X\in\{2D,3D\}$ and Twin-Transformer as follow 
\begin{equation}
    \begin{aligned}
&\operatorname{Transformer}_{TWIN, X}(h_{q}, h_{X}) = h^L_{X}, X\in \{2D, 3D\}\\
&\quad \quad h^l_{2D} = T^l_{2D}(h^{l-1}_{2D}, h_{2D} \| h^{l-1}_{3D}) \\
     &\quad \quad h^l_{3D} = T^l_{3D}(h^{l-1}_{2D}, h_{3D} \| h^{l-1}_{2D})
    \end{aligned}
    \label{eqn:twin}
\end{equation}
where $\|$ is the sequence concatenation operation, $h^l_{2D}, h^l_{3D}$ are the intermediate encoded feature sequences from $l$-th layer in 2D and 3D branch and $\operatorname{Transformer}_{TWIN, X}$ is the Twin-Transformer backbone for 2D or 3D modality.
In Twin-Transformer, the intermediate features from both 2D and 3D modalities are compactly integrated without altering the existing model architecture, thus preserving the 2D VL knowledge through the utilization of feature concatenation and an additional cross-attention mechanism. Additionally, to retain the initial cross-attention between 2D vision and language tokens in the 3D branch, we initialize the 3D branch with the same parameter from 2D branch. By end-to-end finetuning the whole model with 3D-VQA supervision, the crucial 2D VL knowledge could be kept and refined, similar as finetuning 2D VLM for 2D VQA.

\subsection{Question Answering}
Upon acquiring the 2D-3D fused features, we leverage the BLIP's language decoder to generate free-form answers. This approach is particularly advantageous for unseen answer generalization and for datasets requiring long text output, such as ScanQA. The shared language decoder $\operatorname{Transformer}_{VLM,D}$ between 2D and 3D branches receives the fused vision-language feature and computes the word sequence probability as follows:
\begin{equation}
    \begin{aligned}
        p_{2D} &= \operatorname{Transformer}_{VLM,D}(h_{q}, h'_{2D}) \\
        p_{3D} &= \operatorname{Transformer}_{VLM,D}(h_{q}, h'_{3D}) \\
    \end{aligned}
\end{equation}
where $p_{2D}\in [0,1]^N$ and $p_{3D}\in [0,1]^N$ represent the predicted answers of $N$ textual tokens from the 2D and 3D branches, respectively. The entire model is optimized using a perplexity-based loss:
\begin{equation}
    \begin{aligned}
        \mathcal L_{QA}=\sum_i -\lambda \log p_{3D,i,a_i} - \log p_{2D,i,a_i}
    \end{aligned}
\end{equation}
with $\lambda$ as a balancing factor, $(a_1,...,a_N)$ as the answer word index sequence, $p_{X,i,a_i}$ as the predicted probability of token $a_i$ at $i$-th position for modality $X\in\{2D, 3D\}$.
This approach enables joint optimization of the 2D and 3D branches. During inference, the prediction probabilities from both branches are simply summed to generate the final answer. Following BLIP \cite{li2022blip}, we rank the perplexities of answer candidates to narrow down the prediction to the dataset answer-vocabulary.

\begin{table*}[ht]
\centering
\begin{tabularx}{\textwidth}{c|YYY|YY}
\hline
\multirow{2}{*}{Method} & \multicolumn{3}{c|}{ScanQA}        & \multicolumn{2}{c}{SQA} \\ \cline{2-6} 
                        & val   & test w/ obj & test w/o obj & val    & test           \\ \hline
% Scanrefer+MCAN \cite{azuma2022scanqa}        & 18.59 & 20.56       & 19.04        & -      & -              \\
ScanQA w/o multiview \cite{azuma2022scanqa}    & -     & 22.49       & 20.05        & -      & -              \\
ScanQA \cite{azuma2022scanqa}                  & 21.05 & 23.45       & 20.90        & -      & -    \\
SQA \cite{ma2022sqa3d}                  & - & -       & -        & -      & 47.20    \\
FE-3DGQA \cite{zhao2022towards}                & 22.26 & -           & -            & -      & -              \\
CLIP-Guided \cite{parelli2023clip}           & -     & 23.92       & 21.37        & -      & -   \\
Multi-CLIP \cite{delitzas2023multi}             & -     & 24.02       & 21.48        & -      & 48.02  \\
3DVLP \cite{Jin_2023_CVPR}             & 21.65     & 24.58       & 21.56        & -      & -  \\
3DVLP \cite{zhang2023visionlanguage}             & \underline{24.03}     & -       & -        & -      & -  \\
3D-VisTA \cite{zhu20233dvista}             & -      & \underline{27.0}       & \underline{23.0}        & -      & \underline{48.5}  \\
% Ours (2D Only)          & 25.72 &             &              & 50.81  & 51.97          \\
% Ours$^\star$                    & - & 31.19 &  \textbf{30.87}  & - &       \textbf{53.32}           \\
Ours                    & \textbf{26.98} &  \textbf{31.29}(31.19$^\star$)   &    30.82(\textbf{30.87}$^\star$)     & \textbf{52.05}  &       52.91(\textbf{53.32}$^\star$)         \\ \hline

\end{tabularx}
\caption{Comparison of 3D-VQA top-1 accuracy (EM@1) conducted on ScanQA and SQA datasets. Best performance is marked bold and second-best except our method is underlined. $^\star$: Use question to match 2D views at test time. }
\label{tab:main}
\end{table*}

\begin{table*}[ht]
% \fontsize{9}{11}\selectfont
\centering
\begin{tabularx}{\textwidth}{YYYYYYY}
\hline
Method                                    & EM@1           & BLEU-1         & BLEU-4         & ROUGE          & METEOR         & CIDEr          \\ \hline
\multicolumn{7}{l}{\textbf{Test set w/ objects}}                                                                                                \\ \hline
% \multicolumn{1}{c|}{Scanrefer+MCAN\cite{azuma2022scanqa}}       & 20.56          & 27.85          & 7.46           & 30.68          & 11.97          & 57.36          \\
% \multicolumn{1}{c|}{ScanQA w/o multiview\cite{azuma2022scanqa}} & 22.49          & 30.82          & 9.66           & 33.37          & 13.17          & 57.36          \\
\multicolumn{1}{c|}{ScanQA \cite{azuma2022scanqa}}               & 23.45          & 31.56          & 12.04          & 34.34          & 13.55          & 67.29          \\
\multicolumn{1}{c|}{CLIP-Guided \cite{parelli2023clip}}          & 23.92          & 32.72          & 14.64          & 35.15          & 13.94          & 69.53          \\
\multicolumn{1}{c|}{Multi-CLIP \cite{delitzas2023multi}}           & 24.02          & 32.63          & 12.65          & 35.46          & 13.97          & 68.70          \\
\multicolumn{1}{c|}{3DVLP \cite{Jin_2023_CVPR}}                & 24.58          & \underline{33.15}    & 11.23          & 35.97          & 14.16          & 70.18          \\
\multicolumn{1}{c|}{3D-VisTA \cite{zhu20233dvista}}             & \underline{27.0}     & -           & \underline{16.0}     & \underline{38.6}     & \underline{15.2}     & \underline{76.6}     \\
\multicolumn{1}{c|}{Ours}                 & \textbf{31.29} & \textbf{34.49} & \textbf{24.06} & \textbf{43.26} & \textbf{16.51} & \textbf{83.75} \\ \hline
\multicolumn{7}{l}{\textbf{Test set w/o objects}}                                                                                               \\ \hline
% \multicolumn{1}{c|}{Scanrefer+MCAN\cite{azuma2022scanqa}}       & 19.04          & 26.98          & 7.82           & 28.61          & 11.38          & 53.41          \\
% \multicolumn{1}{c|}{ScanQA w/o multiview\cite{azuma2022scanqa}} & 20.05          & 30.84          & 12.80          & 30.60          & 12.66          & 59.95          \\
\multicolumn{1}{c|}{ScanQA \cite{azuma2022scanqa}}               & 20.90          & 30.68          & 10.75          & 31.09          & 12.59          & 60.24          \\
\multicolumn{1}{c|}{CLIP-Guided \cite{parelli2023clip}}          & 21.37          & \underline{32.70}    & 11.73          & 32.41          & 13.28          & 62.83          \\
\multicolumn{1}{c|}{Multi-CLIP \cite{delitzas2023multi}}           & 21.48          & 32.69          & 12.87          & 32.61          & \underline{13.36}    & 63.20          \\
\multicolumn{1}{c|}{3DVLP \cite{Jin_2023_CVPR}}                & 21.56          & 31.48          & \underline{15.84}    & 31.79          & 13.13          & \underline{63.40}    \\
\multicolumn{1}{c|}{3D-VisTA \cite{zhu20233dvista}}             & \underline{23.0}     &  -            & 11.9           & \underline{32.8}     & 12.9           & 62.6           \\
\multicolumn{1}{c|}{Ours}                 & \textbf{30.82} & \textbf{34.41} & \textbf{17.74} & \textbf{41.18} & \textbf{15.60} & \textbf{79.34} \\ \hline
\end{tabularx}
\caption{Comparison of EM@1 performance and text similarity metrics on two test splits of ScanQA. Best performance is marked bold and second-best is underlined. }
\label{tab:main2}
\end{table*}

\begin{table}[ht!]
\fontsize{9}{11}\selectfont
\centering
\begin{tabularx}{\linewidth}{YY|c}
\hline
\multicolumn{2}{c|}{Method}                                     & \multirow{2}{*}{EM@1} \\
Use Answer For Question Transform &  Use Declaration for View Selection &         \\ \hline
       &    &      26.90           \\
$\checkmark$  &    &  25.60          \\
         & $\checkmark$ & \textbf{26.98}                   \\ \hline
\end{tabularx}
\caption{Ablation study on 2D view selection. First line stands for using question for view selection.
Conducted on validation split of ScanQA. Best performance is marked bold. }
\label{tab:ablation-selection}
\end{table}

\section{Experiment}
\label{sec:exp}

\begin{table}[ht!]
% \fontsize{9}{11}\selectfont
\centering
\begin{tabularx}{\linewidth}{YY|Y}
\hline
\multicolumn{2}{c|}{Method}                                     & \multirow{2}{*}{EM@1} \\
3D Input & 2D Input &                                                    \\ \hline
         & $\checkmark$                                  & 25.72                   \\
% $\checkmark$       &                                       & 7.31                    \\/
$\checkmark$       &                                       & 13.99                    \\
$\checkmark$       & $\checkmark$                       & \textbf{26.98}                   \\ \hline
\end{tabularx}
\caption{Ablation study of effectiveness of using 2D and 3D modality.
Conducted on validation split of ScanQA. Best performance is marked bold. 
% /$^*$: 3D branch initialized from scratch.
}
\label{tab:ablation-mod}
\end{table}

\begin{table}[ht!]
\fontsize{9}{11}\selectfont
\centering
\begin{tabularx}{\linewidth}{YYY|Y}
\hline
\multicolumn{3}{c|}{Method}                                     & \multirow{2}{*}{EM@1} \\
 2D-3D Fusion & QA Decoder & Share Decoder                        \\ \hline
% 2D Only       & $\checkmark$    & $\checkmark$     & 25.72    \\
Concat       & $\checkmark$     & $\checkmark$    & 25.90                   \\
Simple       & $\checkmark$     & $\checkmark$    & 26.45                   \\
Twin         &           & $\checkmark$ & 21.37                   \\
Twin         & $\checkmark$     &     &   26.49          \\
Twin         & $\checkmark$     & $\checkmark$    & \textbf{26.98}                   \\ \hline
\end{tabularx}
\caption{Ablation study of different 2D-3D fusion mechanism based on 2D VLM, and the method design for the question decoder. Conducted on validation split of ScanQA. Best performance is marked bold. }
\label{tab:ablation-fusion}
\end{table}

\subsection{Training and Evaluation}

\noindent \textbf{Dataset and Evaluation:} 
We evaluate our model on two key 3D VQA datasets: ScanQA, with 41K question-answer annotations across 800 indoor 3D scenes from ScanNet (including two test sets with or without ground-truth object annotation), and SQA, with 33K question-answer annotations also based on 650 scenes in ScanNet. Following previous work \cite{parelli2023clip,azuma2022scanqa}, we assess 3D-VQA performances using top-1 accuracy (EM@1) for both datasets. For ScanQA, where answers are often long-text and free-form, we additionally report text similarity metrics (BLEU, ROUGE, METEOR, and CIDEr scores), as conventionally adopted in previous work. The SQA dataset come with an additional asker location annotation and a ``situation" text describing the asker location, enabling potential embodied AI applications.

\noindent \textbf{Model Initialization and Training Setup:} 
To effectively transfer vision-language knowledge from 2D to 3D contexts, we initialize our 2D-3D VQA models using a pretrained 2D VLM, namely BLIP. The image encoder, 2D-VL encoder, and answer decoder in the 2D branch are BLIP-initialized, while the 3D-VL encoder is also initialized from the 2D-VL encoder for initial modality alignment. On both the ScanQA and SQA datasets, training is conducted for 10 epochs using AdamW optimizer, with learning rates of 1e-5/3e-5 for the 2D/3D branch.

\noindent \textbf{Dataset Specific Implemetation Details:}
Following ScanQA \cite{azuma2022scanqa} and SQA \cite{ma2022sqa3d}, we employ auxiliary tasks to assist the 3D VQA task that are conventionally adopted on two datasets. Specifically, the 3D object detector backbone from VoteNet is jointly supervised using scene object detection loss $\mathcal L_{det}$. For ScanQA, we incorporate question-related object localization $\mathcal L_{objloc}$ and classification losses $\mathcal L_{objcls}$, resulting in a total loss term for ScanQA of $\mathcal L=\mathcal L_{QA}+\mathcal L_{det}+\mathcal L_{objloc}+ \mathcal L_{objcls}$. In the case of SQA, we utilize a localization loss $\mathcal L_{subloc}$ to regress the subjective location, with the total loss term for SQA being $\mathcal L=\mathcal L_{QA}+\mathcal L_{det}+\mathcal L_{subloc}$. The additional predictions are made through lightweight MLP heads following the fused 2D-3D features. 
For SQA dataset, we process the additional ``situation" description by encoding the text $t_s$, into situation representations $h_{s} = \operatorname{Enc}(t_s)$, and concatenating it with the 3D object tokens, $h_{3D} \leftarrow h_{3D} \| h_s$, where $\|$ denotes sequence concatenation, $\operatorname{Enc}$ is the text encoder and $h_{3D}$ is the encoded 3D object features. 
% /The main methodology remains largely unchanged.

\subsection{Results}

\textbf{Effectiveness of \methodNameLong{}:} To evaluate the efficacy of \methodNameLong{}, we juxtapose our approach with prior 3D-VQA methods (Table \ref{tab:main} and \ref{tab:main2}). Table \ref{tab:main} contrasts the top-1 accuracy (EM@1) on both ScanQA and SQA datasets, while Table \ref{tab:main2} additionally measures text-similarity metrics on two ScanQA test splits, thereby more precisely assessing the ability to predict long-text and free-form answers. Among the baselines compared in Table \ref{tab:main}, ScanQA (w/o multiview) and FE-3DGQA (Line 1,4) are 3D-only methods that utilize a 3D detector to extract object features and predict answers through a Transformer-based QA module. ScanQA also explored (Line 2) using multiview point cloud feature projected from all corresponding 2D view features. CLIP-Guided and Multi-CLIP (Line 5,6) incorporate pretrained 2D VLM and 2D views, but in the way of global feature alignment with top-down 2D view. 
We also compare our method with a series of initial work on 3D VL pretraining \cite{Jin_2023_CVPR, zhang2023visionlanguage, zhu20233dvista} (Line 7-9), that first pretrain the backbone with different pretraining tasks and finetune on 3D VQA.

Our proposed method has yielded significant advances on both datasets, leading by margins of 3.0\%, 4.3\%, and 7.8\% compared to the second-best performances on the validation, test w/ objects, and test w/o objects splits of the ScanQA dataset, respectively, along with margins of 4.4\% on the test split of the SQA dataset. And without 3D vision-language pretraining (3D VLP), we surpassed these 3D VLP methods (Table \ref{tab:main}, Line 7-9) by introducing efficient 2D-3D knowledge transfer and fusion procedure.
Furthermore, since the exactly-match accuracy is not the only standard for VQA involving long-text and free-form answers,  our approach also exhibits a significant performance gain evaluated by text similarity metrics (Table \ref{tab:main2}). 
Collectively, these results empirically validate the effectiveness of \methodNameLong{}, against previous state-of-the art methods.

We tested 3D VQA performance without question-to-declaration transformation at test time, to reduce additional inference cost. The results were similar, with variations of -0.1\%, +0.05\%, and +0.41\% across three test splits, still maintaining a significant lead over other methods. Thus, the transformation is optional at test time and can be optionally omitted for quicker inference.

\subsection{Ablation Study}
\label{sec:ablation}

% \yang{swap the table 3 and table idx, you should always discuss smaller id Table in the main text}
\textbf{Choice of View Selection Condition:}
We evaluated the best text for selecting question-conditional views, finding a 0.1\% performance gain using declarations over questions (Table \ref{tab:ablation-selection}, Line 1, 3). Although this result verifies the benefit of using declarations, it also shows that questions are already effective with image-text retrievers like BLIP. We further examined using both question and answer to convert questions into declarations during training. While it enhanced 2D view selection, it introduced a domain shift between training and evaluation, resulting in a 1.5\% performance decrease (Table \ref{tab:ablation-selection}, Line 2, 3), empirically demonstrates the consequent domain-shift incorporating answer. In conclusion, our design choice that using question-to-declaration transform during training achieved the best perforamance.
\begin{figure}[t]
    \centering
    \includegraphics[width=\columnwidth]{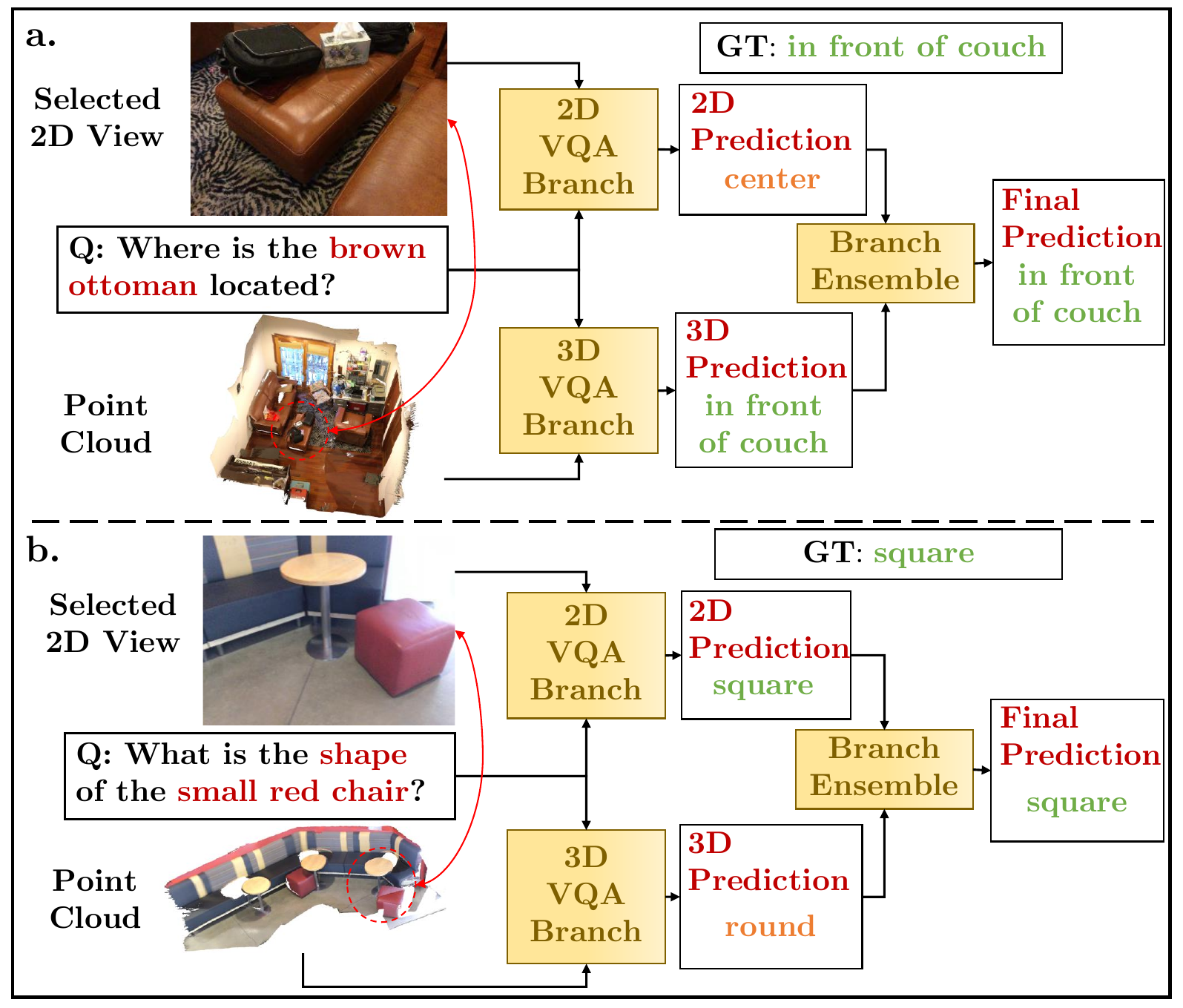}
    \caption{Qualitative results on how the 2D and 3D branch mutually help each other to correct the answer. The dotted circle in 3D scene corresponds to the 2D view.}
    \label{fig:qual}
\end{figure}

\noindent \textbf{Effectiveness of Using 2D and 3D Visual Input:}
We investigated the impact of 2D and 3D visual modality inputs in our proposed method (Table \ref{tab:ablation-mod}). Our multi-modal approach outperformed the uni-modal methods utilizing solely 2D or 3D visual input, achieving a margin of 1.3\% and 13.0\% 
respectively (Line 3 vs. Line 1, 2). These findings underscore the challenges associated with simply substituting 2D visual tokens with 3D visual tokens in 3D VQA, as this can detrimentally affect performance due to the visual feature shift between the 2D and 3D encoders. Furthermore, the results empirically validate the advantage of harnessing both 2D and 3D modalities, demonstrating improvement in 3D VQA performance as compared to a 2D-only approach. In conclusion, both 2D and 3D visual context should be used to enhance 3D-VQA performance.

\noindent \textbf{Effectiveness of Introducing Twin-Transformer:}
We assessed the efficacy of our Twin-Transformer 2D-3D fusion mechanism by comparing it with several baseline methods on the validation split of ScanQA, as detailed in Table \ref{tab:ablation-fusion}. Here, ``Twin" denotes our proposed Twin-Transformer, ``Concat" stands for the approach that employs a single-stream Transformer to accept both 2D and 3D visual tokens by simply concatenating the sequences from the two modalities as input, and ``Simple" refers to the structure that mirrors the Twin-Transformer but without the intermediate feature cross-attention mechanism. The Twin-Transformer's 3D-VQA performance surpassed the single-stream concatenation and naïve two-branch training by margins of 1.1\% and 0.5\%, respectively (Line 2, 3, 6). These results underscore the effectiveness of our proposed Twin-Transformer mechanism in efficiently aligning and fusing 2D and 3D visual contexts.

\noindent \textbf{Ablation for Question Answering Language Decoder:}
To ablate the design in on utilizing the pretrained question answering language decoder, we conduct experiments as illustrated in Table \ref{tab:ablation-fusion}. When substituting the language decoder with an MLP-based classification head, the 3D-VQA performance declined by a substantial margin of 5.6\% (Line 4, 6). This outcome underscores the significance of retaining the pretrained language knowledge from the 2D VLM to predict answers.
Using a shared decoder achieved a performance increase of 0.5\% (Line 5, 6). This result empirically confirms the efficiency of the decoder sharing strategy in enabling the answer decoder to comprehend both 2D and 3D fused vision-language features.

\subsection{Qualitative Result}
In our qualitative results (Figure \ref{fig:qual}), we highlight two instances of our model's dual-branch predictions: (a) The selected 2D views are insufficient to answer the question, as they only partially show the couch. However, thanks to 3D visual clues from the other branch, which encodes all 3D objects, the model answers correctly; (b) The 3D branch fails to distinguish the attributes ``round" and ``square" due to limited scene data, but a well-chosen view, aided by pretrained 2D visual language understanding, allows the 2D branch to differentiate the two attributes, correcting the model's prediction. These examples further confirm the two branches' ability to mutually enhance 3D VQA performance.

\section{Conclusion}
\label{sec:conclusion}

In this paper, we introduced \methodNameLong{}, a novel approach for training a 3D Visual Question Answering (VQA) model that leverages 2D pretrained vision-language models to capture both 2D and 3D visual clues. By selecting question-related 2D views, this method is able to harness potential 2D visual clues to enhance the understanding of 3D scenes. The proposed two-branch Twin-Transformer architecture allows for the seamless fusion and alignment of 2D and 3D modalities, leading to improved 3D VQA performance. Validated on the ScanQA and SQA datasets, our method has demonstrated effectiveness in 3D VQA, accommodating both 2D and 3D visual contexts, and surpassing current state-of-the-art methods.

\section{Acknowledgements}
\label{sec:ack}
This work was supported by the grants from the National Natural Science Foundation of China 62372014.

\bibliography{aaai24}

\end{document}